# Fast Graph Construction Using Auction Algorithm

Jun Wang and Yinglong Xia
IBM T. J. Watson Research Center
1101 Kitchawan Rd, Yorktown Heights, NY 10598, USA
{wangjun, yxia}@us.ibm.com

## Abstract

In practical machine learning systems, graph based data representation has been widely used in various learning paradigms, ranging from unsupervised clustering to supervised classification. Besides those applications with natural graph or network structure data, such as social network analysis and relational learning, many other applications often involve a critical step in converting data vectors to an adjacency graph. In particular, a sparse subgraph extracted from the original graph is often required due to both theoretic and practical needs. Previous study clearly shows that the performance of different learning algorithms, e.g., clustering and classification, benefits from such sparse subgraphs with balanced node connectivity. However, the existing graph construction methods are either computationally expensive or with unsatisfactory performance. In this paper, we utilize a scalable method called auction algorithm and its parallel extension to recover a sparse yet nearly balanced subgraph with significantly reduced computational cost. Empirical study and comparison with the state-of-art approaches clearly demonstrate the superiority of the proposed method in both efficiency and accuracy.

## 1 INTRODUCTION

Graph based data representation treats data points as graph nodes and builds pairwise edges between these nodes which are often weighted by the similarity or correlations between the corresponding data samples. In the past decade, graph based data representation has become very popular and widely used in many machine learning algorithms due to both practical needs and theoretical advantage. First, under many practical situations, the data come in a natural way of graph representation, where the links or connections between items can be treated as directed or undirected edges. Typical examples include social networks, the Internet, and citations networks [13, 21]. Second, for the conventional vector space based data representation, a data graph can be easily generated using pre-defined similarity or correlation measurement.

Given the popularity of data graph representation, many graph based machine learning algorithms have been developed, including graph based clustering and semi-supervised learning algorithms. For instance, representative graph based clustering approaches include those formulated as a graph cut problem [16, 20]. In addition, graph has also been integrated into semi-supervised learning paradigm, resulting in a wide range of approaches, such as graph Laplacian regularization framework [2, 23, 25] and random walk based methods [1, 22]. Other popular methods include graph based ranking algorithms [17, 24]. In many real applications and systems, graph based approaches have shown empirical success, especially for those with agnostic setting and given no prior knowledge of the data distributions.

As indicated in [12, 15], constructing a suitable data graph is the vital step for ensuring the learning performance. The sparse structure of the constructed graph is desired for learning algorithms to remain efficiency to memory usage and robustness to noisy samples. Given an adjacency graph, a typical way to recover a sparse graph without losing the global structure of the data is to build a neighborhood graph. The objective is to connect each data point with its nearest neighbors and prune all other connections. However, since the nearest neighborhood method prune the edges through a greedy process, it can only produce an asymmetric graph with some adhoc symmetrization processes. As a result, the constructed neighborhood graph is often irregular and imbalanced, where the learning performance could be significantly degraded [12]. Therefore, this drawback of nearest neighbor method motivated one to consider matching technique, such as $b$-matching, to build a balanced graph structure [11, 12]. Instead of achieving symmetrization by post-processing, $b$-matching sets the symmetry as additional constraints and directly extracts a symmetric solution. But this modification imposes much high computation and memory cost and the one of the known fast approximated solution using belief propagation has the space complexity of $O(n)$ and the time complexity $O(n^{2.5})$ [10], where $n$ is the number of graph nodes (refer to Table 1).

Considering the trade off between computational efficiency and performance, in this paper, we propose a fast and scalable solution for constructing nearly balanced graphs using the auction algorithm proposed in [3]. In order to remain performance superiority and reduce computation cost, our method tends to extract an approximately balanced graph though preserving the regularity as much as possible. We show that this method has the space complexity $O(n)$ and time complexity $O(|\mathbf{E}| \cdot \max |a_{ij}|/\epsilon)$, where $\max |a_{ij}|/\epsilon$ is a constant determined by the data and the learning parameter $\epsilon$. It is much less expensive than the existing regular graph construction methods, e.g. $b$-matching technique. In addition, the auction algorithm based graph construction is naturally parallelizable and therein the computational efficiency can be further improved using multiple processes. We provide extensive experiments and comparison study with the state-of-the-arts on both synthetic and real benchmark datasets. The experimental results clearly show that the graphs created by parallel auction algorithm (PAA) provide superior performance than neighborhood graphs for different machine learning algorithms. Although PAA graphs have slightly lower quality than $b$-matching graphs, we observe that PAA method reduces the computational cost by hundreds of times than $b$-matching method. Therefore, the PAA method is very suitable for handling large scale applications with massive data graphs.

The remainder of this paper is organized as follows. In Section 2, we give a brief introduction of related work on graph construction process. A new graph construction and sparsification method using a parallel auction algorithm is presented in Section 3. The experiments are reported in Section 4 followed by conclusions and discussions in Section 5.

## 2 RELATED WORKS

Although graph based machine learning methods have been extensively studied and explored, there have been limited efforts for building effective data graphs in the community. Although some recent methods use sampling techniques to approximate large scale graphs [14], it still remains as a challenging issue to efficiently construct large scale graphs with good qualities, such as sparsity and regularity. In this section, we will provide a brief introduction about related work for graph construction and then motivate our proposed method.

Given a data set $\mathbf{X}$, graph construction aims in converting the data to a weighted graph $G = \{\mathbf{X}, \mathbf{E}, \mathbf{W}\}$ where graph nodes are the samples and $\mathbf{W} \in \mathbb{R}^{n \times n}$ is a weight matrix of edges $\mathbf{E}$. As summarized in [12], two important steps for graph constructions are sparsifying the adjacency matrix $\mathbf{A} = \{a_{ij}\} \in \mathbb{R}^{n \times n}$ and assigning weights to edges. In many practical settings, the adjacency matrix is computed using a similarity function for all pairs of nodes, i.e. $a_{ij} = sim(\mathbf{x}_i, \mathbf{x}_j)$. For other applications, such as social network analysis and relational learning, the adjacency or connection matrix $\mathbf{A}$ might be given in a natural way without the above process of computing pair-wise similarity. In the second step, graph sparsification tries to recover a sparse binary matrix $\mathbf{B} = \{b_{ij}\} \in \mathbb{B}^{n \times n}$ to preserve the global structure of data using local information. The final edge matrix can be easily derived as $\mathbf{W} = \mathbf{B} \bullet \mathbf{A}$, where the symbol $\bullet$ indicates element-wise multiplication. Such an extract sparse subgraph often leads better learning performance and requires less storage and computation cost [11, 12].

The most popular method for constructing a sparse graph is the nearest neighbor (NN) approach, including different variants such as $k$-nearest neighbor and $\epsilon$-nearest neighbor methods. Since $\epsilon$-nearest neighbor often generate fragile subgraphs and does not provide satisfactory performance, $k$NN graph remains more popular in many applications [15]. Briefly speaking, $k$NN graph construction performs a greedy search process to select the edges with the most significant weights for each node, and remove all the other insignificant ones. The objective of $k$NN approach indeed solves the following maximization problem

$$\max_{b_{ij}} \sum_i \sum_j b_{ij} a_{ij}$$
$$\text{s.t.} \quad \sum_j b_{ij} = k, \qquad (1)$$
$$b_{ii} = 0, b_{ij} \in \{0,1\}, \forall i,j = 1, \cdots, n \quad .$$

Since the greedy search for $k$ nearest neighbors does not guarantee to produce symmetric $\mathbf{B}$, it usually needs some postprocessing to achieve symmetrization by $\max(\mathbf{B}, \mathbf{B}^\top)$ or $\min(\mathbf{B}, \mathbf{B}^\top)$. However, this adhoc symmetrization often leads to imbalanced connectivity for nodes, which could seriously degrade the learning

Table 1: Comparison of the time and space complexity for different graph construction approaches. Assume the input adjacency graph has $n$ nodes and $|\mathbf{E}|$ edges. The compared methods are: Loopy Belief Propagation (LBP) [9], Sufficient Selection Belief Propagation (SSBP) [10], and the auction algorithm.

| Method | Time | Space |
|---|---|---|
| LBP Method | $O(n|\mathbf{E}|)$ | $O(n+|\mathbf{E}|)$ |
| SSBP Method | $O(n^{2.5})$ | $O(n)$ |
| Nearest Neighbor Method | $O(n^2)$ | $O(1)$ |
| Auction Method | $O(|\mathbf{E}| \cdot \max |a_{ij}|/\epsilon)$ | $O(n)$ |

performance, as observed in [12].

As a valuable alternative for extracting sparse graph structure, weighted $b$-matching technique [18] has been applied to build graph with balanced node connectivity. Similarly to the objective of $k$NN approach, $b$-matching recovers a sparse graph with the maximized edge weights. Furthermore, $b$-matching enforces the symmetry of connectivity through directly imposing a set of constraints $b_{ij} = b_{ji}$. Essentially, $b$-matching is a generalized version of the well-known linear assignment problem, which can be solved by Hungarian method with a cubic complexity $O(n^3)$ ($n = |\mathbf{X}|$) in time. Various methods have been proposed to solve $b$-matching problem with less memory and time cost, including the reduction based method [8] and belief propagation based approaches [9, 10], for which the computational complexity is summarized in Table 1. Although $b$-matching method yields clear advantage for different machine learning algorithms, including spectral clustering and semi-supervised classification [11, 12], it is a much more expensive procedure than the intuitive $k$NN method. Due to both time and memory constraints, the existing $b$-matching based graph construction is infeasible for large scale datasets.

## 3 AUCTION ALGORITHM FOR GRAPH CONSTRUCTION

In this section, we describe the procedure to utilize an auction algorithm to identify $bn$ significant edges from the given bipartite graph corresponding to the adjacency matrix $\mathbf{A}$. In particular, each node is connected with at most $b$ edges. If given an arbitrary graph, it is straightforward to convert it to a bipartite graph though creating a shadow node for each graph node and only connect the original graph nodes to the shadow nodes. In order to utilize distributed computing systems to process massive data graphs, we also describe the parallelization technique for the proposed graph construction method.

### 3.1 Auction Algorithm for Linear Assignment

Here, we first describe the solution for a single-edge matching problem using an auction algorithm, and then provide general extensions. For a standard maximum 1-matching problem (also known as a linear assignment problem) over a bipartite graph with a nonnegative weight matrix $\mathbf{A}$, the objective can be written as:

$$\max_{\mathbf{B}} \; \text{tr}(\mathbf{A}^\top \mathbf{B}) \qquad (2)$$
$$\text{s.t.} \quad \mathbf{B}\vec{1}^T = \vec{1}, \; \mathbf{B}^T \vec{1} = \vec{1}, \; \mathbf{B} \in \mathbb{B}^{n \times n},$$

where $\vec{1} = (1,1,\cdots,1)^T$. This linear program can be interpreted as selecting a permutation matrix $\mathbf{B}$, so that after permutation of $\mathbf{A}$, the sum of entries on the diagonal is maximized. The edges corresponding to the permutation matrix $\mathbf{B}$ is as known as a matching in $\mathbf{A}$. Thus, the solution of this program can be used to sparsify $\mathbf{A}$ with constraints of regularity. To generalize the standard linear assignment problem to the objective of extracting a balanced graph where each node connects to $b$ edges, we can simply replace $\vec{1}$ with $\vec{b} = b \cdot \vec{1}$. However, this solution for such a $b$-matching problem can be approximated by solving the above linear assignment problem for $b/2$ times and removing the selected edges after each iteration.

As discussed earlier, we seek fast and scalable solutions for graph sparsification due to the demand of processing large scale graphs in many domains. Conventional approach for solving Eq. 2 is based on increasing augmenting paths, which leads to low concurrency and limited scalability [3, 19]. Thus, we propose to use the auction algorithm introduced by Bertsekas in [3]. Since the constrained problem in Eq. 2 can be converted into a dual form, the auction algorithm deals with the above linear programming by solving its dual problem. In particular, we are solving the following dual problem of the original primal linear program:

$$\min_{p,\pi} \; \vec{1}^T p + \vec{1}^T \pi \qquad (3)$$
$$\text{s.t.} \quad \vec{1}p^T + \pi\vec{1}^T \geq \mathbf{A},$$

where $p = (p_1, p_2, \cdots, p_n)^T$ and $\pi = (\pi_1, \pi_2, \cdots, \pi_n)^T$ are known as the dual vectors. The constraints $\vec{1}p^T + \pi\vec{1}^T \geq \mathbf{A}$ indicate the element wise relationship between the matrices $\vec{1}p^T + \pi\vec{1}^T$ and $\mathbf{A}$, i.e. $(\vec{1}p^T + \pi\vec{1}^T)_{ij} \geq a_{ij}, i, j = 1, \cdots, n$. The auction algorithm computes the dual variables through an iterative *auction* process. Intuitively, we view each row $i$ as a buyer and each column $j$ as an object; $a_{ij}$ is the benefit of object $j$ to buyer $i$; $p_j$ is the price of object $j$ (initialized to 0 at the beginning); $\pi_i$ is the profit for buyer $i$, defined as $(a_{ij} - p_j)$, in case object $j$ is assigned to $i$. The auction algorithm proceeds as follows: Each buyer $i$ bids for an adjacent object $j$ that offers the highest profit, i.e., $\max_{j \in \text{adj}(i)}(a_{ij} - p_j)$, where $\text{adj}(i)$ denotes the set of adjacent nodes of $i$. The bid is the difference between the highest profit and the second highest profit. An object $j$ is assigned to the buyer offering the highest bid, say buyer $i$. Thus, row $i$ and column $j$ are matched and price $p_j$ is increased by the bid. Price increment makes an object expensive for competitors, so that they can choose other objects. The above steps are repeated for all unmatched rows until every row is matched or the matching does not change.

The naive auction algorithm described above has a defect caused by *price war* [3]. When the highest and second highest profits are equal for some buyers, the bid becomes 0 and the object price remains unchanged. Therefore, buyers can indefinitely compete for the same object without increasing its price. The object will be alternately assigned to them and the auction will not progress. To overcome this issue, a small positive scalar $\epsilon$ is added to the price of an object once it is assigned to a buyer. A smaller $\epsilon$ leads to better accuracy; a greater $\epsilon$ results in an accelerated auction process.

### 3.2 Auction for Graph Sparsification

The auction algorithm described in Section 3.1 selects a single edge for each node while maximizing the total weight of the selected edges. For selecting $b$ edges, there are two straightforward approaches to generalize the auction. The first approach is to perform auction multiple times. Every time the algorithm selects an edge for each node, if exists, without having two edges connect to the same node. The other approach is to select up to $b$ edges in each iteration process, following the same criteria for edge selection. The details are discussed in below.

Note that the above algorithm does not ensure that if a node $i$ connects to a shadow node $j'$, node $j$ will also connect to node $i'$. This can cause adverse impact on graph regularity. We propose a heuristic strategy to reduce such negative impact. When such inconsistence

---

**Algorithm 1** Fast Graph Construction Using Parallel Auction Algorithm

**Graph partitioning**: partition graph G into $L$ disjoint subgraphs $\{G_l\}_{l=1}^L$ with weight matrices $\{\mathbf{A}_l\}_{l=1}^L$, where $\mathbf{A}_l = \{a_{ij}^l\}$

**Initialization:** Dual vectors $p = \vec{0}$, initial $\pi = \vec{0}$, the set of selected edges $\mathcal{M} = \emptyset$

**while** price keeps changing **do**

[1]. For each subgraph corresponding to the weight matrix $\mathbf{A}_l = \{a_{ij}^l\}$, compute for nodes in $X_l$ that have not identified $b$ edges:

$$j_1^l = \arg\max(a_{ij}^l - p_j^l), \forall j \in \text{adj}(i)$$

......

$$j_{b+1}^l = \arg\max(a_{ij}^l - p_j^l), \forall j \in \text{adj}(i), j \neq j_1, \cdots j_b$$

[2]. Remove earlier selected edges that conflict

$$\mathcal{M} = \mathcal{M} \setminus \{(i', j_t^l) \in \mathcal{M}, \forall 1 \leq t \leq b, i' \in X_l\}$$

[3]. Add current selected edges

$$\mathcal{M} = \mathcal{M} \cup \{(i, j_t^l) \in \mathcal{M}, \forall 1 \leq t \leq b\}$$

[4]. Update price

$$p_{j_t^l}^l = a_{ij_t^l}^l - a_{ij_{t+1}^l}^l + p_{j_{t+1}^l}^l + \epsilon, \forall t = 1, \cdots, b$$

[5]. Synchronize price globally $p_j^l = \max_{t=1}^P p_j^t$

**end while**

---

occurs, saying node $j$ selects node $\tilde{i}'$ instead of node $i'$, we compare the percentile of edge $(j, \tilde{i}')$ with respect to all incident edges of $j$ and the percentile of $(i, j')$ with respect to all incident edges of $i$. By percentile, we mean the index of the selected edge among all incident edges sorted in descending order. We accept the edge with smaller percentile while reject the other one. The intuition that we select edges according to the percentile instead of edge weight is that all incident edges for the boundary nodes in a graph can be weighed much higher than even the lightest incident edge of an internal node.

### 3.3 Parallelization

In this subsection, we discuss a parallelization strategy of the auction algorithm for graph sparsification. Given the weight matrix $\mathbf{A}$ of a bipartite graph, we first partition $\mathbf{A}$ into $L$ disjoint sub-matrices, where each sub-matrix $\mathbf{A}_l = \{a_{ij}^l\}$ is of size $n_l \times n$, $\sum_{l=1}^L n_l = n$. Therefore, sub-matrix $\mathbf{A}_l$ is the weight matrix of a bipartite subgraph $G_l$, which contains disjoint node sets $X_l$ and $Y_l$. Although $\mathbf{A}$ can also be partitioned by columns or into checkerboard structures [4], the above horizontal partitioning along row direction is more natural since large scale sparse graphs/matrices are often stored in the compress sparse rows format. After partitioning, we can rewrite the linear assignment problem in Eq. 2 as:

$$\max \sum_{l=1}^{L} \sum_{i \in X_l} \sum_{j \in Y_l} a_{ij} b_{ij} \qquad (4)$$

$$\text{s.t.} \sum_{j \in Y_l} b_{ij} = 1, \ \forall i \in X_l, l = 1, \cdots, L$$

$$\sum_{l=1}^{L} \sum_{i \in X_l} b_{ij} = 1, \ \forall j \in Y_l, \quad b_{ij} \geq 0, \ \forall i, j.$$

Table 2: Average running time (seconds) on the synthetic bipartite and unipartite graphs using different methods for constructing sparse graphs. The graphs have nodes from 100 to 2000 and the value of $b$ is set as 10 uniformly. The compared methods include Loopy Belief Propagation (LBP) [9], $k$NN, and the proposed parallel auction algorithm (PAA).

| $n$ | Bipartite | | | | Unipartite | | | |
|---|---|---|---|---|---|---|---|---|
| | $|\mathbf{E}|$ | LBP | $k$NN | PAA | $|\mathbf{E}|$ | LBP | $k$NN | PAA |
| 100 | $5 \times 10^3$ | 0.2 | $8 \times 10^{-4}$ | $4 \times 10^{-3}$ | $9.9 \times 10^3$ | 0.3 | $4 \times 10^{-3}$ | $1.1 \times 10^{-2}$ |
| 200 | $2 \times 10^4$ | 2.8 | $3 \times 10^{-3}$ | $1.5 \times 10^{-2}$ | $3.98 \times 10^4$ | 0.7 | 0.01 | 0.04 |
| 500 | $1.25 \times 10^5$ | 18.7 | 0.02 | 0.11 | $2.495 \times 10^5$ | 6.2 | 0.07 | 0.28 |
| 1000 | $5 \times 10^5$ | 37.9 | 0.11 | 0.53 | $9.99 \times 10^5$ | 23.2 | 0.39 | 1.26 |
| 2000 | $2 \times 10^6$ | 609.7 | 0.47 | 2.42 | $3.998 \times 10^6$ | 111.4 | 1.31 | 5.39 |

Eq. 4 implies a straightforward parallelization of auction algorithm for graph sparsification. In Algorithm chart 1, we summarize the major steps of parallel auction for graph sparsification. Here we set the initial values for $\vec{p}$ and $\vec{\pi}$ are both $\vec{0}$. The maximum number of edges allowed for a node is denoted by $b$ and the set for selected edges is $\mathcal{M}$, which is initialized as an empty set $\emptyset$.

The above algorithm allows concurrent auction on a maximum of $L$ processors, each of which handles one single bipartite subgraph. The only communication between processors occurs when synchronizing price vectors in step 5 of the above parallel algorithm. Note that the solution of the other dual variable $\vec{\pi}$ is updated implicitly, which is given by $\pi_i^l = a_{ij}^l - p_j^l$, where $(i, j)$ is selected. The parallel auction typically accelerates computation due to the concurrent auction activities in each sub-matrix. However, it does not change the complexity. That is, in the worst case (the input graph is a chain), it is equivalent to a serial auction process.

### 3.4 Complexity Analysis

The time complexity for auction algorithm depends on the number of auction iterations and the workload for each iteration. In an auction process, a price of node increases at least for $\epsilon$ [3], i.e., the minimum bid increment. Since the profit for bidder $i$ that matched with object $j$ is given by $(a_{ij} - p_j)$, the profit becomes negative after $|a_{ij}|/\epsilon$ iterations, i.e., $i$ will seek other objects for possibly higher profit. In each iteration, we visit at most $|\mathbf{E}|$ edges (the number typically decreases dramatically after a few iterations), so the overall complexity is $O(|\mathbf{E}| \max |a_{ij}|/\epsilon)$. During the auction process, the only information we must keep is the price of objects. There are $n$ objects, so the space complexity is $O(n)$. Note that the space for storing the data or graph itself is not counted for evaluating space complexity. In Table 1, we summarize the time and space complexity of the proposed parallel auction algorithm based graph construction process and compare with other methods. From this analysis, it is easy to see the proposed parallel auction algorithm (PAA) based method is extremely efficient for large scale sparse graph data, which is indeed very common in real applications, such as social network and geometry analysis [7].

## 4 EXPERIMENTS

In this section, we provide empirical study of the proposed parallel auction algorithm (PAA) based graph construction method and compare with the-state-of-art approaches. In particular, we are interested in comparing with the belief propagation (BP) based methods [9, 10] since these method tends to generate high quality graphs with balanced node connectivity, similar as our proposed method. Note that there are two variants of the BP approach, loopy belief propagation (LBP) and a speed-up version, sufficient selection belief propagation (SSBP). However, in the empirical study, we notice that these two methods share very similar computational performance. Hence, we only report the results of LBP for comparison. In addition, we also implement the nearest neighbor search based graph construction using the standard sorting pro-cess as another baseline. Extensive experimenters over both synthetic and real benchmark datasets are performed to evaluate the computational efficiency and performance for different learning algorithms, as discussed below.

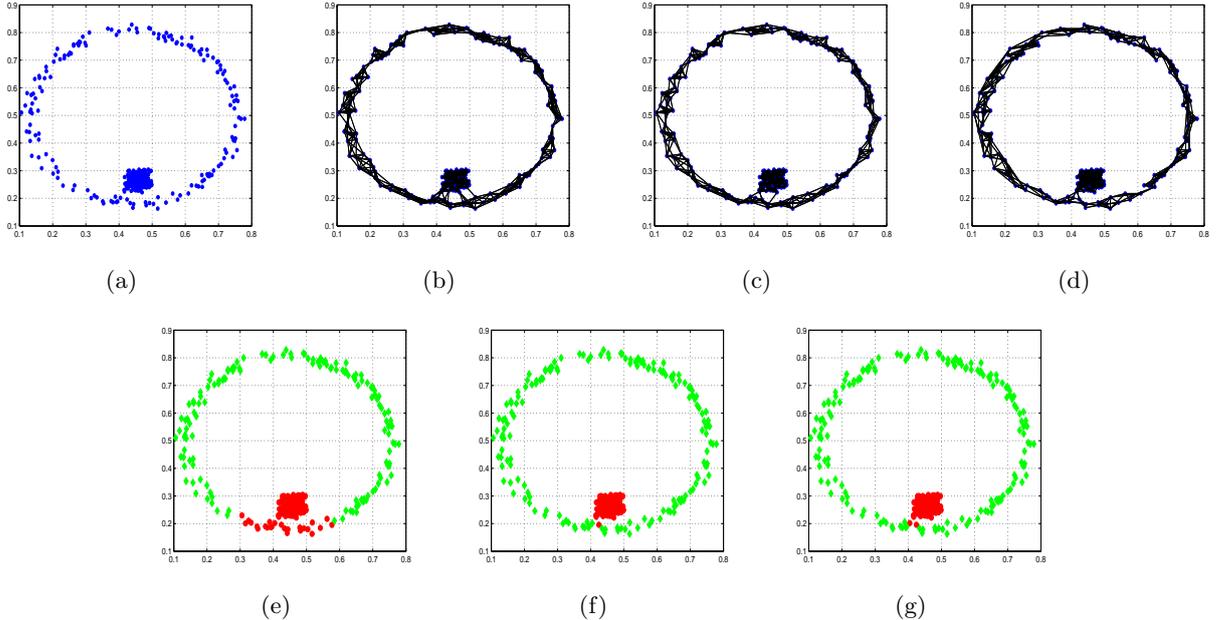

Figure 1: Illustration of the constructed graphs and clustering results over the toy datasets using different methods: a) original data; b) $k$-nearest neighbor graph ($k = 8$); c) $b$-matching graph ($b = 8$); d) PAA graph. Different color indicate the extract clusters using: e) $k$NN graph; f) $b$-matching graph; g) PAA graph.

### 4.1 Speed Comparison

We first evaluate the running time of different graph construction methods. Similar to the settings in [9], here we randomly generate bipartite and unipartite graphs with $100 \leq n \leq 2000$ nodes, where the weights of graph edges are independently sampled from a uniform distribution $U(0, 1)$. For bipartite graph, each of the two disjoint node sets has $n/2$ nodes. For constructing a sparse graph, the degree of each nodes of the extracted sparse graph is uniformly set for all the experiments. For instance, in $k$NN approach, the value of $k$ is set as 10, and accordingly $b = 10$ for $b$-matching methods. For fair comparison, all the experiments are conducted on the same platform with identical hardware and software environment.

For the LBP method, we use the default parameters in the software package [1], except setting the maximum iteration as 5000. Table 2 shows the comparison of the running time for the synthetic data. It is obvious that $k$NN provides the most efficient graph construction method due to its fast greedy search process. However, comparing LBP and PAA, both of which tend to generate balanced graphs, PAA improves the speed of graph construction in the magnitude of $10^2$. In addition, we found that the propose method has more stable running performance for different random data, while both LBP method relies more on the properties of the synthetic graph data.

---
[1] http://www.cs.umd.edu/b̃ert/code/bmatching/

### 4.2 Performance Comparison

Graph representation has been widely used for different machine learning algorithms, including clustering and classification. In this subsection, we present the performance comparison study of clustering and classification through using different graph construction approaches. For clustering analysis, spectral clustering has been developed and shown effective in practice [16]. In a standard spectral clustering process, there are three key steps. The first step is to construct a data graph that connects sample points. Then one has to perform eigen-decomposition over the graph Laplacian. Finally, K-means algorithm is applied to derive the final clusters. In this study, we use the toy data [12] to construct different graphs, namely $k$NN graph, $b$-matching graph, and the graph built by the parallel auction algorithm (PAA-graph). Figure 4(a)-1(d) show the original data and different graphs. Then we feed these graph into standard spectral clustering algorithm to compare the results.

Figure 2 shows the error rate for the clustering results. The number of connected edges for graph nodes varies from 6 to 24. For all the tests, we use the same Gaussian kernel based weighting scheme with the kernel bandwidth informally set as 0.1. Clearly, $k$NN graph provides the worst performance and $b$-matching graph has the best performance. The proposed parallel auction based graph construction method has slightly worse performance than $b$-matching. However, it significantly increases the by the magnitude of $10^2$ in practice. Figure 1(e)-1(g) shows the examples of constructed sparse graphs using different approaches. Due to the greedy search process, $k$NN method generates highly irregular graph and breaks the data structure by creating more cross-manifold edges. In contrast, $b$-matching and the proposed PAA method create more balanced graphs and have much less edges across the two clusters. When the node degree increases, it is easy to generate more cross-cluster edges. Hence, all the clustering performance of all three graph construction methods accordingly decreases.

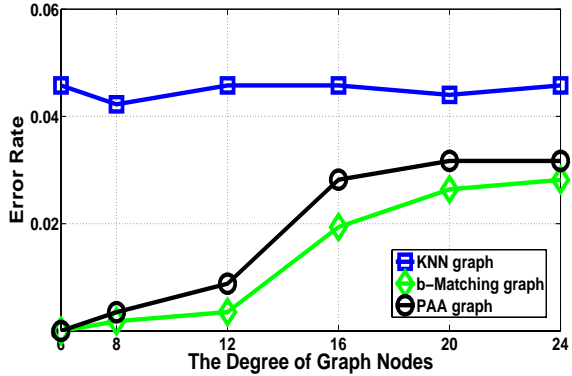

Figure 2: The performance comparison of spectral clustering using different graphs. The horizontal axis indicates the number of connected edges for graph nodes and the vertical axis shows the error rate of clustering results.

In addition, we also compare the performance of graph based semi-supervised learning for classification tasks, i.e. classifying USPS handwritten digits [5] and face recognition [6]. In particular, we apply the method developed by Zhou et.al in [23] as the testbed for different graph construction methods. In the tests, we fix the number of node degree as 7 and use the same weighting scheme, similar to the settings used in the above clustering experiments. The number of initially given labels is varied from 4 to 20 for digits data and 3 to 18 for the face data. For each setting, we randomly run 50 independent trials and compute the average error rates of the classification results. Figure 3(a) and 3(b) show the performance curves with error bars for digits classification and face recognition, respectively. Again, $b$-matching approach builds the most balanced graph and therein achieves the best performance. The proposed PAA method has slightly worse performance but almost as fast as $k$NN method, which tends to create the highly imbalanced graphs.

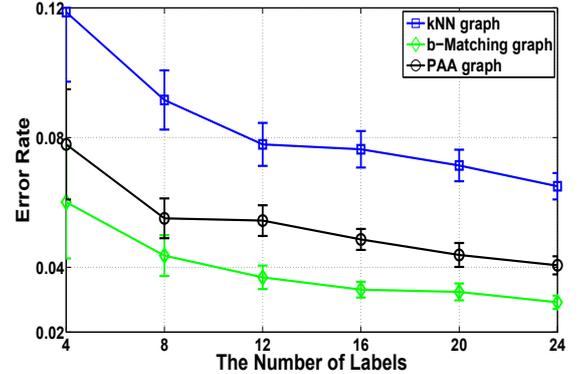

(a)

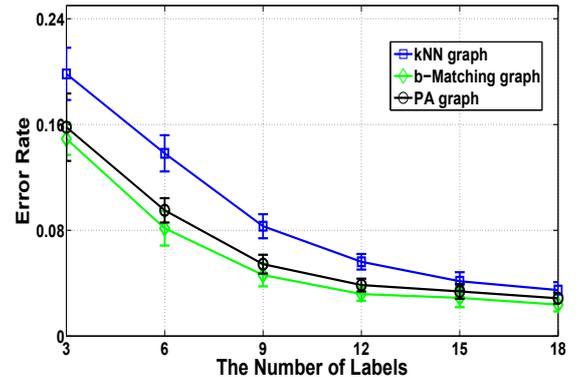

(b)

Figure 3: The performance comparison of graph based semi-supervised learning method using different graph construction methods: a) digits classification; b) face recognition. The horizontal axis indicates the number of initially given labels and the vertical axis shows the error rate of classification results.

### 4.3 Evaluation of Parallel Efficiency

Since all the above experiments involve the graphs with relatively small scales, there is no extra benefit to parallelize the algorithm. However, when handling large scale data, it is very important to perform parallel process for fast graph construction. Note that the time complexity of the PAA based graph construction is linear with respect to the number of edges. This property makes the approach very suitable to the applications with large scale sparse data graphs. In order to validate the parallelization efficiency of the proposed method for such applications, here we used two real benchmark datasets about circuit simulation problems from the sparse matrix collection [7]. The first graph data "ASIC_680ks" contains $682,712$ nodes and $2,329,176$ edges (after reverse process) [2], and the second data "ASIC_680k" has $682,862$ nodes and $3,871,773$ edges (after reverse process) [3]. Figure 4 presents the visualization of these two large graphs[4]. From this figure, it is easy to see that both graphs are heavily irregular and the boundary nodes are with extremely sparse connectivity. Here we apply the proposed PAA method to recover sparse and balanced graphs from the original graph data.

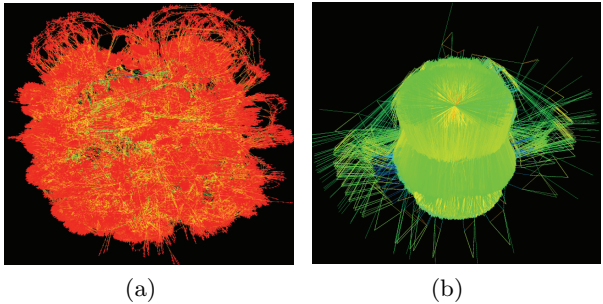

(a)　　　　　　　(b)

Figure 4: The visualization of the two large graph data: a) ASIC_680ks, and b) ASIC_680k.

Our experiments are performed using a computer cluster on which each compute node contains two 2.33 GHz Intel Xeon E5410 quadcore processors. The eight cores share 32 GB DDR2 RAM running at 800 MHz. The OS is Ubuntu 9.10 with GCC 4.10 installed. All codes were compiled with -O3 level option. We spawn parallel processes using MPICH2 to evaluate the parallel performance of the proposed PAA based graph construction method. In addition, we utilize MPICH2 so that our proposed technique can scale up to multiple compute nodes when the scale of the input graph is large enough. The graphs are partitioned evenly and launched to each process with equal number of nodes along with all the neighbor information.

For both graph data, graph construction is performed for two scenarios: $b = 2$ and $b = 4$. For those nodes with less than $b$ edges in the original graph, all the connected edges will be preserved. We vary the number of used processes from 1 to 8, where each process is hosted by a separate core. The computational cost includes the time for generating parallel process, partitioning the graph, assigning tasks to each process, and performing the auction process (The time for loading data is not considered). Table 3 provides the running time in seconds for each tested case. From this table, it is easy to see that the proposed PAA-based graph construction method achieves lower execution time when multiple processes were used. However, the speedup is *sublinear* instead of linear because the extra time used for parallelization and communication. In general, when performing the graph construction in 8 parallel processes, the time cost will be reduced from one quarter to more than half. Especially for more complex problems (e.g., $b = 4$), the speed up is usually more significant.

---

[2] http://www.cise.ufl.edu/research/sparse/matrices/Sandia/ASIC_680ks.html
[3] http://www.cise.ufl.edu/research/sparse/matrices/Sandia/ASIC_680k.html
[4] The visualization examples are available from the website: http://www.cise.ufl.edu/research/sparse/matrices/

Table 3: Evaluation of the parallel efficiency of the proposed PAA-based graph construction method. The numbers give the computational cost in seconds using different number of processes.

| Data | ASIC_680ks | | ASIC_680k | |
|---|---|---|---|---|
| # of processes | $b=2$ | $b=4$ | $b=2$ | $b=4$ |
| 1 | 3.09 | 18.92 | 1.76 | 11.62 |
| 2 | 2.74 | 18.73 | 1.49 | 10.83 |
| 4 | 2.70 | 11.60 | 1.33 | 7.29 |
| 8 | 2.47 | 8.92 | 1.26 | 5.03 |

## 5 CONCLUSION

Data graph has been widely used for different machine learning and data mining algorithms. It is critical to build high quality graphs with affordable cost. Commonly, one aims in extracting sparse yet regular graph from data for both theoretic and practical advantages. Popular methods for constructing sparse graphs include nearest neighbor method and $b$-matching algorithm. The former is very efficient since it essentially perform greedy search to pick up the significant graph edges. However, it often generates highly imbalanced graphs with unsatisfactory performance. The later can provide fairly balanced regular graphs but suffers from extremely high computational cost.

In order to remain computational efficiency, while achieving certain properties of the extracted graph, such as sparsity and regularity, in this paper, we proposed a novel graph construction method through exploring the auction algorithm. In particular, the proposed method is almost as fast as $k$NN method, while maintain stratificatory regularity property. Experimental results show that the parallel auction algorithm based graph construction provides comparable performance to the $b$-matching algorithm, but improves the speed in the magnitude of $10^2$. Finally, we also provide the empirical study of the parallelization efficiency over large data sets with up to 680K nodes. One of our future direction is to extend the proposed graph construction to semi-supervised and supervised scenarios, where partial label information are given to help improve accuracy of the auction process.

## Acknowledgements

This research was partially supported by the CRA/CCC Computing Innovation Fellow (CIFellow) Award. We would like to thank Dr. Anshul Gupta in the IBM T.J. Watson Research Center for the discussion on parallelization of the auction algorithm.